\documentclass[conference]{IEEEtran}

\usepackage{subfigure}
\usepackage{algorithm, algorithmic}
\usepackage{multicol,multirow}
\usepackage{caption}
\usepackage{graphicx}
\usepackage{amsmath,amssymb,bbold}
\usepackage{booktabs}
\usepackage{hyperref}
\hypersetup{
    colorlinks=true,
    linkcolor=black,
    citecolor=black,
    filecolor=black,      
    urlcolor=black,
    }
\usepackage{cleveref}
\usepackage{placeins}
\graphicspath{{./plots/}}
\newcommand{\Prob}{\text{I\kern-0.15em P}}
\def\NDCG{\mathrm{NDCG}}
\def\DCG{\mathrm{DCG}}
\def\IDCG{\mathrm{IDCG}}

\newcommand{\permSpace}{\mathbb S}
\newtheorem{definition}{Definition}
\begin{document}
%
\title{Fairness in Ranking: Robustness through Randomization without the Protected Attribute}

\author{
\IEEEauthorblockN{Andrii Kliachkin}
\IEEEauthorblockA{University of Padova}
\IEEEauthorblockN{Eleni Psaroudaki}
\IEEEauthorblockA{NTUA \& ARC}
\and
\IEEEauthorblockN{Jakub Mareček}
\IEEEauthorblockA{CTU}
\IEEEauthorblockN{Dimitris Fotakis}
\IEEEauthorblockA{NTUA \& ARC}
}

\author{
\IEEEauthorblockN{Andrii Kliachkin\IEEEauthorrefmark{1},
Eleni Psaroudaki\IEEEauthorrefmark{2}\IEEEauthorrefmark{3},
Jakub Mare\v{c}ek\IEEEauthorrefmark{4}, and 
Dimitris Fotakis\IEEEauthorrefmark{2}\IEEEauthorrefmark{3}
}

\IEEEauthorblockA{\IEEEauthorrefmark{1}
Department of Mathematics, Università degli Studi di Padova, Padua, Italy.}
\IEEEauthorblockA{\IEEEauthorrefmark{2}
School of Electrical \& Computer Engineering, National Technical University of Athens, Greece.}
\IEEEauthorblockA{\IEEEauthorrefmark{3} 
Athena Research \& Innovation Center in Information Communication \& Knowledge Technologies, Athens, Greece.}
\IEEEauthorblockA{\IEEEauthorrefmark{4}The Department of Computer Science, Czech Technical University in Prague, the Czech Republic.}
}


%


\maketitle

\begin{abstract}
There has been great interest in fairness in machine learning, especially in relation to classification problems. 
In ranking-related problems, such as in online advertising, recommender systems, and HR automation, much work on fairness remains to be done.
Two complications arise: first, the protected attribute may not be available in many applications. 
Second, there are multiple measures of fairness of rankings, and optimization-based methods utilizing a single measure of fairness of rankings may produce
rankings that are unfair with respect to other measures. 
In this work, we propose a randomized method for post-processing rankings, which do not require the availability of the protected attribute. 
%
%
In an extensive numerical study, we show the robustness of our methods
with respect to P-Fairness and effectiveness with respect to Normalized Discounted Cumulative Gain (NDCG) from the baseline ranking, improving on previously proposed methods. 
\end{abstract}


\IEEEpeerreviewmaketitle

\section{Introduction}

There has been a great interest in fairness in machine learning \cite[e.g.]{barocas2023fairness} over the past decade 
With the coming regulation of AI both in the European Union and the US, the interest is expected to grow substantially.  
Much of the work on fairness in machine learning has focused on fairness in classification, so far.

Beyond classification, there are first papers on fairness in forecasting \cite[e.g.]{zhou2023fairness}, decision making \cite[e.g.]{jabbari2017fairness}, and ranking \cite[cf.]{10.1145/3533379,10.1145/3533380}.
In some ways, one may argue that ranking and decision-making have a much larger impact on humans than most classification problems.  
Consider, for example, ranking within HR automation: a recruiter may obtain hundreds of applications for a job, 
and needs to shortlist 10 best candidates for the hiring manager to interview. This task is increasingly being automated 
(cf. workable.com, linkedin.com), and the ranking algorithms utilized in the online platforms need to be fair. 
(For the legal perspective, see the AI Act, Annex III, fifth paragraph.)
Similarly, recommender algorithms in social media increasingly drive political preferences, 
and their performance is of paramount importance for the continued existence of political systems of many countries. 
Many other applications arise in information retrieval \cite{liu2009learning}.


Two complications arise. First, there are multiple measures of fairness of rankings. Proportionate fairness (P-Fairness) may be the best known, but many others are possible.  Optimization-based methods for post-processing or aggregating rankings often utilize only a single measure of the fairness of rankings, which may produce rankings that are not unfair with respect to other measures. We call this the challenge of robustness. Moreover, achieving fairness may compromise efficiency, which is usually quantified by normalized discounted cumulative gain (NDCG) or distance from a baseline ranking. 

Second, the protected attribute may not be available in many applications.
Consider, for example, the HR-automation use case, as above. 
There, neither the hard-copy resumes, nor the profiles on online platforms such  Workable or LinkedIn, 
contain protected attributes such as gender, religion, or ethnicity.
In many legal systems (e.g., France), it is illegal to collect such information.  
At the same time, both the employer and the vendor of the AI system
can be fined, if they are found guilty of direct or indirect discrimination. 
Following pioneering work on uncertainty in the group membership \cite{awasthi2020equalized,wang2020robust}, 
fairness without demographics has attracted much recent attention in classification
\cite[e.g.]{chai2022fairness,zhao2022towards,han2023retiring,zhou2023group},
but has not been studied within fairness in ranking, yet. 

Here, we propose a randomized method for postprocessing rankings to improve their fairness.
Inspired by approaches of differential privacy \cite{dwork2014algorithmic}, where noise is admixed to data to protect privacy, 
the proposed method admixes Mallow's noise to improve fairness. 
Thus, our method does not require the availability of the protected attribute to improve fairness of rankings 
in a way oblivious to the specific protected attributes. 
At the same time, our method addresses the challenge of robustness. 
In an extensive numerical study, we show the robustness of our method
with respect to P-Fairness 
and competitive efficiency with respect to Normalized Discounted Cumulative Gain (NDCG) and 
KT distance from the baseline ranking,
improving upon previously proposed methods
ApproxMultiValuedIPF \cite{dong2022rankaggregation} and DetConstSort \cite{Geyik_2019}, 
where the latter has been developed by a team at LinkedIn.

Overall, our contributions are:
\begin{itemize}
\item a randomized method for post-processing of rankings improving fairness
\item a computational study of five algorithms evaluated with respect to P-Fairness and multiple protected attributes and with respect to two different measures of efficiency.
\end{itemize}

\section{Related Work} 
\label{related} 

Geyik et al. \cite{Geyik_2019} introduced the problem of computing efficient rankings under proportionate fairness constraints \cite{BaruahCPV96,CelisSV18}. Proportionate fairness (or P-fairness, in short) requires that each protected group is represented by at least a given proportion of individuals in each prefix of the final ranking. On the other hand, efficiency requires that the final ranking is close to an initial quality-optimal ranking, where the individuals appear in non-increasing order of their quality-based scores, without any fairness or other considerations. In \cite{Geyik_2019}, efficiency is quantified by NDCG. 
Geyik et al. \cite{Geyik_2019} presented and experimentally evaluated a few natural heuristics (including {\sc DetConstSort}, which we compare against in this work) that aim to balance between violating P-fairness and maximizing NDCG. 

Subsequently, Wei et al. \cite{dong2022rankaggregation} and Chakraborty et al. \cite{ChakrabortyD0S22} presented efficient algorithms for computing P-fair rankings, under a stricter definition of proportionate fairness using both upper and lower bounds on the representation of each protected group, where efficiency is measured with respect to the distance of the final ranking to a given initial ranking. They proved that the best P-fair ranking can be computed in polynomial time for the Kendall tau distance with two protected groups (Algorithm {\sc GrBinaryIPF}, inspired by mergesort, in \cite{dong2022rankaggregation}) and with any number of protected groups (Theorem~3.4 in \cite{ChakrabortyD0S22}), for the Spearman's footrule distance with any number of protected groups (Algorithm {\sc ApproxMultiValuedIPF}, based on the computation of a minimum weight bipartite matching, in \cite{dong2022rankaggregation}) and for the Ulam distance with a constant number of protected (Theorem~3.10 in \cite{ChakrabortyD0S22}). 

Wei et al. \cite{dong2022rankaggregation} and Chakraborty et al. \cite{ChakrabortyD0S22} also considered the more general problem of aggregating a collection of rankings to a P-fair ranking that minimizes the total distance to the given ones. In a nutshell, their approach is to aggregate the given rankings into a near-optimal ranking with respect to the objective of minimizing the total distance, which is a well studied problem \cite{AilonCN08,Kenyon-MathieuS07} from the viewpoint of polynomial-time approximation algorithms, and then to transform that ranking into a P-fair one using the algorithms above. 

An orthogonal direction, extensively studied in computational social choice (see e.g., \cite{Kempe2024} and the references therein), is to aggregate a collection of rankings, submitted by a set of voters, into a final ranking under proportional representation constraints for any group of sufficiently large and sufficiently cohesive group of voters. For an excellent discussion of a substantial volume of work in this direction under different notions of proportional representation and different objectives, we refer the reader to \cite[Section~6]{Kempe2024}. 

In this work, we propose and experimentally evaluate the use of Mallows noise \cite{mallows1957non} in order to produce approximately efficient rankings, starting from a given one and considering the objectives of maximizing NDCG and minimizing the Kendall tau distance, which are also approximately P-fair with respect not only to some given protected subgroups, but also to other (possibly unknown) protected subgroups that are sufficiently large. The Mallows model has received considerable attention from the viewpoints of both modelling and quantifying noise in computational learning tasks that involve rankings, e.g., \cite{fotakis2021aggregating,fotakis2022linear,Psaroudaki2022}, and of computationally and statistically efficient learning of Mallows distributions \cite{busa2019optimal,liu2022finding}. To the best of our knowledge, our work is the first that investigates the use of Mallows noise towards achieving P-fairness with respect to groups possibly defined by unknown protected attributes.

\section{Technical Background}
\subsection{Preliminaries}
Let a set $C$ of $d$ of candidates or items. We denote the symmetric group over $d$ elements with $\permSpace_d$ and, for incomplete rankings, $\permSpace_{\leq d}$. Let $\sigma \in \permSpace_d$ be a \textbf{ranking} (or permutation) of the $k$ elements. For $i \in [d]$, we denote $\sigma(i)$ as the position of the $i$-th element. Each candidate $i \in C$ has a\textbf{ protected attribute} $\mathcal{A}(i)$, that can take any of $g$ values, leading to  $g$ \textbf{protected groups} of candidates. If $g=2$ then we have a binary protected attribute, leading to 2 groups $G_1$ and $G_2$, while if $g>2$ we have a multi-valued protected attribute leading to more groups $G_1, \dots, G_g$.

\subsection{Fairness Metrics for Rankings}
We want to satisfy some fairness metrics in the suggested ranking. A class of fairness measures, usually referred to as P-fairness, corresponds to the proportional representation of each protected group in the top-$k$ or in every prefix of the top-$k$ of the ranking. The following definitions formalize two different variants of P-fair rankings, namely $(\vec{\alpha}, \vec{\beta})$-$k$ fair rankings and $(\vec{\alpha}, \vec{\beta})$-weak $k$-fair rankings.

\begin{definition}[($\vec{\alpha}, \vec{\beta}$)-$k$ fair ranking, Def. 2.4 of \cite{ChakrabortyD0S22}]\label{def:pfairness}
Consider a set $C$ of $d$ candidates partitioned into $g$ groups $G_1, \dots, G_g$ and $\vec{\alpha}=(\alpha_1, \dots, \alpha_g) \in [0,1]^g$, $\vec{\beta}=(\beta_1, \dots, \beta_g) \in [0,1]^g$, $k \in [d]$. A ranking $\pi \in \permSpace_d$ is said to be $(\vec{\alpha}, \vec{\beta})$-$k$ fair if for any prefix $P$  of $\pi$ of length at least $k$ and each group $i \in [g]$, there are at least $ \lfloor \alpha_i \cdot |P| \rfloor  $ and at most $ \lceil  \beta_i \cdot |P| \rceil$ elements from the group $G_i$ in $P$, i.e.,
\[\forall_{prefix} \,{P: |P|\geq k}, \forall{i \in [g]}, \lfloor \alpha_i \cdot |P| \rfloor \geq |P \cap G_i|  \geq \lceil  \beta_i \cdot |P| \rceil\]
\end{definition}

\begin{definition}[($\vec{\alpha}, \vec{\beta}$)-weak $k$-fair ranking, Def. 2.5 of \cite{ChakrabortyD0S22}]\label{def:weak-pfairness}
Consider a set $C$ of $d$ candidates partitioned into g groups $G_1, \dots, G_g$ and $\vec{\alpha}=(\alpha_1, \dots, \alpha_g) \in [0,1]^g$, $\vec{\beta}=(\beta_1, \dots, \beta_g) \in [0,1]^g$, $k \in [d]$. A ranking $\pi \in \permSpace_d$ is said to be $(\vec{\alpha}, \vec{\beta})$-weak $k$-fair if for the $k$-length prefix $P$ of $\pi$ and each group $i \in [g]$, there are at least $ \lfloor \alpha_i \cdot k \rfloor  $ and at most $ \lceil  \beta_i \cdot k \rceil$ elements from the group $G_i$ in $P$, i.e.,
\[ \forall{i \in [g]}, \lfloor \alpha_i \cdot k \rfloor \geq |P \cap G_i|  \geq \lceil  \beta_i \cdot k \rceil\]
\end{definition}

A quantitative notion of fairness in rankings, introduced in \cite{dong2022rankaggregation} and referred to as Infeasible Index, accounts for  percentage of positions satisfying P-fairness. 

\begin{definition}[Two-Sided Infeasible Index] 
\begin{align*}
    \text{LowerViol}(\pi) & = \sum_{k=1}^{|\pi|} \mathbb{1}{(\exists G_i \in G, ~s.t. ~ \text{count}_k(G_i, \pi) < \lfloor \beta_i \cdot k\rfloor} \\
\text{UpperViol}(\pi) &= \sum_{k =1}^{|\pi|} \mathbb{1}{(\exists G_i \in G, ~s.t. ~ \text{count}_k(G_i, \pi) > \lceil \alpha_i \cdot k\rceil}\,
\end{align*}
where $\text{count}_k(G_i, \pi)$ is the number of elements of group $G_i$ in the top $k$ positions of ranking $\pi$. Then,  %
\[
\text{TwoSidedInfInd}(\pi) = \text{LowerViol}(\pi) + \text{UpperViol}(\pi) 
\]
\end{definition}

\begin{definition} [Percentage of P-Fair Positions]
    \[\text{PPfair}(\pi, i) = 100*(1-\text{TwoSidedInfInd}(\pi, i)/|\pi|)\]
\end{definition}

\subsection{Distance Metrics in Rankings} \label{sec:d_metrics}

There are several well-known metrics for measuring the similarity between two rankings $\sigma, \pi\in \permSpace_k$, including Spearman distance, Kendall Tau, Kendall's tau coefficient, etc. We define them all for completeness but focus our reporting on the Kendall Tau distance.

\noindent
\textbf{Spearman distance}  measures the total element-wise displacement of $\pi$ from $\sigma$:
\begin{equation*}\label{eq:spearman}
d_2(\pi, \sigma) = \sum_{i \in [k]} (\pi(i) - \sigma(i))^2   
\end{equation*}

\noindent\textbf{Kendall Tau (KT) distance}  measures the total number of discordant pairs between the two permutations:
\begin{equation*}\label{eq:kt}
d_{KT}(\pi, \sigma) = \sum_{i < j} \mathbb{1}\{ (\pi(i) - \pi(j))(\sigma(i) - \sigma(j)) < 0 \}
\end{equation*}

\noindent\textbf{Kendall's tau coefficient} is the normalization of $d_{KT}$ to the interval $[-1, 1]$:
\begin{equation*}\label{eq:ktc}
    k_\tau = 1 - \frac{4d_{KT}(\pi, \sigma)}{k (k-1)}
\end{equation*}
Kendall's tau coefficient measures the proportion of the concordant pairs between two rankings. The $k_\tau$ is 1 when perfect agreement between the two rankings occurs (i.e.,
the two rankings are the same), and negative when the agreement is less than expected by chance, with -1 when the two rankings have a perfect disagreement.

\subsection{Ranking Quality Measures} \label{sec:q_metrics}
In recommendation engines, a model predicts the rank of a list of items based on the search queries. There are relevance scores for each item, and the quality of a ranking is determined by comparing the relevance of the items to the relevance of the items in the optimal, in terms of quality, ranking. Hence, they aim to maximize a quality measure, such as Cumulative Gain (CG), Discounted Cumulative Gain (DCG), Ideal Discounted Cumulative Gain (IDCG), and \emph{Normalized Discounted Cumulative Gain} (NDCG). In this paper, following the work of \cite{Geyik_2019} we utilize the NDCG metric to assess the quality of the suggested rankings, which is defined as: 
%
\[ \NDCG(\pi) = \frac{\DCG(\pi)}{\IDCG(\pi)} \]
Specifically, $\DCG$ is the sum of gains discounted by rank, computed as :  
\[ \DCG(\pi) = \sum_{i=1}^k \frac{s(\pi(i))}{\log(1+i)}, \] 
while $\IDCG$  calculates the DCG of the ideal order based on the gains, i.e. provides the quality of the optimal ranking $\pi^*$
\[ \IDCG(\pi) = \sum_{i=1}^k \frac{s(\pi^*(i))}{\log(1+i)}. \] 

The optimal ranking $\pi^\ast$ is to list the items in decreasing order of their scores. For simplicity, we can say that since the $\DCG$ of the optimal ranking  (i.e., the $\IDCG$) is known, we focus on the optimization of the $\DCG$.

\subsection{Mallows Model} \label{sec:mm}

We utilize a distance-based probability model introduced by Mallows \cite{mallows1957non}. The standard Mallows model $\mathcal{M}(\pi_0, \theta) $ is a two-parameter model with the probability mass function for a permutation $\pi$ to be equal to:

\[ 
\Prob_{\pi \sim \mathcal{M}(\pi_0, \theta)}[\pi| \pi_0, \theta] = \frac {e^{-\theta d(\pi, \pi_0)}}{Z_k(\theta, \pi_0)} 
\]
$\pi_0$ is the \textit{central ranking} (also known as location parameter) while $\theta$ is the \textit{dispersion parameter} (also known as spread).
The probability of any other permutation exponentially decreases as the distance from the central permutation increases, while the dispersion parameter controls how fast this occurs.  
$Z_k(\theta, \pi_0) = \sum_{\pi \in \permSpace}{e^{-\theta d(\pi, \pi_0)}}$ is the \textit{normalization constant}, which when $d=d_{KT}$, only depends on the dispersion $\theta$ and $k$  and not on the central ranking, hence $Z_k(\theta, \pi_0)= Z_k(\theta)$. 

\subsection{Problem Definition} 
\label{sec:model}

Given a ranking $\pi^\ast \in \permSpace_d$ (which is supposed to have been optimized based on some quality-based scores that may be unknown to our algorithms) and $g$ protected groups $G_1, \ldots, G_g$ with proportionate fairness requirements 
$\alpha$ and $\vec{\beta}$ as in definitions~\ref{def:pfairness}~and~\ref{def:weak-pfairness}, we aim to compute a (possibly approximately and weakly) P-fair ranking $\pi$ which is as close to $\pi^\ast$ as possible. If the quality-based scores behind $\pi^\ast$ are unknown, we use the Kendal tau distance to quantify the efficiency of $\pi$, as in \cite{dong2022rankaggregation,ChakrabortyD0S22}. Otherwise, we use focus on maximizing $\DCG$ (and $\NDCG$) subject to proportionate fairness constraints (as formalized by the Integer Linear Program in Section~\ref{sec:ilp}. 

In Section~\ref{sec:experiments}, we also evaluate the final ranking with respect to proportionate fairness against unknown protected groups, showing that the use of Mallows noise results in robustness of fairness against unknown protected attributes at the expense of reasonable deterioration of the efficiency objective.

\section{Our Algorithms}


\subsection{Post-hoc fairness through Mallows noise} \label{mallows}
In this work, we utilize to the Mallows model (see \Cref{sec:mm}) as a randomization mechanism that ensures fairness in a way oblivious to the groups. 
Specifically, given a central permutation and a dispersion parameter, we sample rankings from the corresponding Mallows distribution, and keep the best according to a specific metric. 

The central ranking could be either the result of a rank aggregation problem or any ranking in general. 
For the problem of fair rankings, we pick a weakly fair ranking of candidates ordered by their descending score as the center of our Mallows distribution.

\begin{algorithm}[tb!]
\caption{Fair ranking algorithm through Mallows noise}
\label{algo1}
\small
\begin{algorithmic}[1]
    \STATE \textbf{Input:} set of items to rank $S$, number of samples $m$, criterion $c$
    \STATE \textbf{Output:} a ranking $\pi$
    \STATE \texttt{noisy\_ranking($m$)}:
    \begin{ALC@g}
    \STATE $\pi_0 \leftarrow$ \texttt{find\_central\_permutation(S)}
    \FOR{$i \in \{1, \dots, m\}$} 
        \STATE $samples[i] \leftarrow \mathcal{M}(\theta, \pi_0)$
    \ENDFOR
    \STATE  $best\_ranking \leftarrow  $ \texttt{choose\_ranking($c, samples$)}
    \RETURN $best\_ranking $
    \end{ALC@g}
\end{algorithmic}
\end{algorithm}






\subsection{Integer Linear Programm} \label{sec:ilp}

The optimal $\vec{\alpha}, \vec{\beta}$)-$k$ fair ranking with respect to the metrics of $\DCG$ and $\NDCG$ can be computed via the following Integer Linear Program (ILP). 

\begin{align*}
\max & \sum_{i \in [k]} \sum_{j \in [k]} s(i) c(j) x_{ij} &  \\
 \mbox{s.t.} & \sum_{i \in [k]} x_{ij}  = 1 & \forall j \in [k] \\
 & \sum_{j \in [k]} x_{ij}  \leq 1 & \forall i \in [k]\\
 & \lfloor \beta_p \ell \rfloor \leq \sum_{i \in G_p} \sum_{j = 1}^\ell x_{ij}  \leq \lceil \alpha_p \ell \rceil & \forall \ell \in [k], \forall p \in [g]
\\
%
&  x_{ij} \in \{ 0, 1 \}
\end{align*}
where for the $\DCG$ and $\NDCG$ metrics, we use $c(j) = 1/\log(1+j)$.


\section{Experimental Evaluation}
\label{sec:experiments}
We utilize Mallow's model as a randomization mechanism that ensures fairness in a way oblivious to groups, without a big loss in terms of optimality. Specifically, we perform three types of experiments, two with synthetic data (\Cref{sec:mallows-ii-exp,sec:uniform-exp}) and one with the German Credit dataset \cite{german_credit_statlog} (\Cref{sec:german-credit-exp}). The code for the reproduction of our results can be found in \url{https://github.com/andrewklayk/fairness_with_mallows_distribution}.
\subsection{Mallow's model and Infeasible Index}
\label{sec:mallows-ii-exp}
The first experiment aims to evaluate the impact of Mallow's randomization on the Infeasible Index of a ranking. The experimental setup and results follow.

\subsubsection{Experimental Setup}
 We analyze a scenario of ten individuals, who belong to two equal-sized groups and create multiple rankings, denoted as $\sigma_{II}$, by adjusting the placement of candidates from each group to produce diverse values of the Infeasible Index (II). We then sample rankings from a Mallows distribution using the $\sigma_{II}$ ranking as the central permutation and varying values of $\theta$, observing the resulting Infeasible Index in the samples. 

\subsubsection{Experimental Results} 
\begin{figure*}[tb!]
  \centering
  \includegraphics[width=0.8\textwidth]{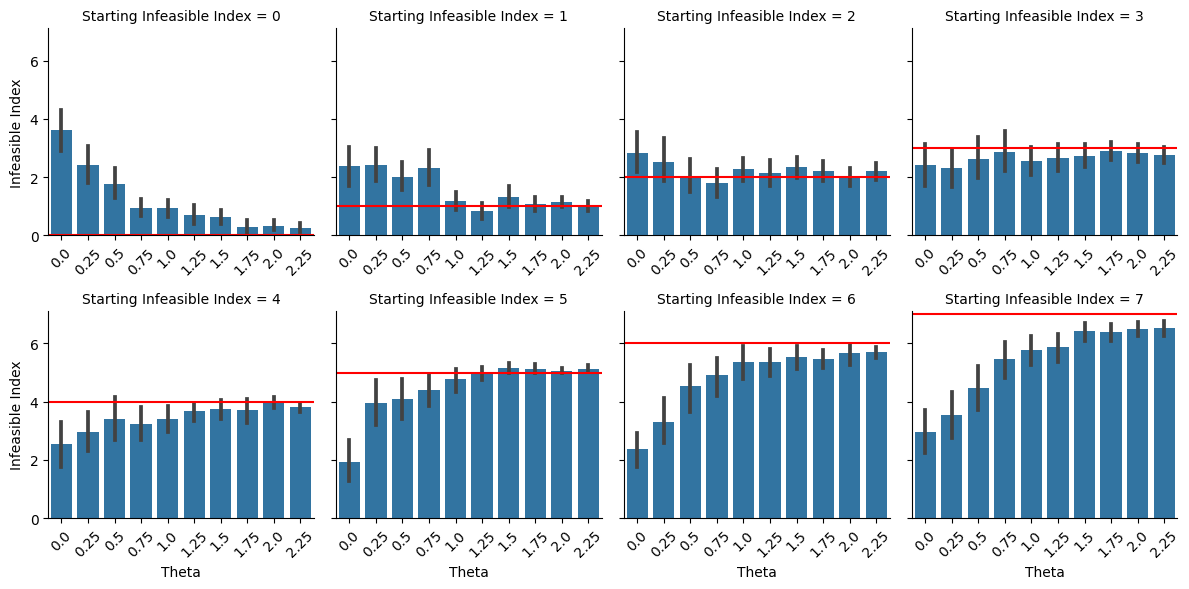}
  \caption{Mallow's distribution and Infeasible Index. Each subplot corresponds to a different value of the central's ranking Infeasible Index. The Infeasible Index of the central ranking is shown as a red line. The bar plots depict the mean value of the Infeasible Index of the samples from the Mallows distribution centered on the initial ranking with two groups. Confidence intervals were obtained via bootstrapping ($n=1000$).}
  \label{plot:experiment1}
\end{figure*}

In Fig.~\ref{plot:experiment1} we observe that as the dispersion parameter increases, the Infeasible Index of samples drawn from Mallow's model converges to the Infeasible Index of the central permutation. When the centrals' permutation Infeasible Index is small, as  $\theta \rightarrow 0$,  we notice that the Infeasible Index of the samples drawn from Mallows' model is higher, but without a significant difference. However, when the centrals' permutation Infeasible Index is large, we notice a significant drop in the Infeasible Index of the samples drawn from Mallows' model as $\theta \rightarrow 0$.

\subsection{Mallow's model and NDCG}
\label{sec:uniform-exp}
The aim of the second experiment is to evaluate the impact of Mallow's randomization on both fairness and ranking utility in a simple ranking setup.

\subsubsection{Experimental Setup}
We consider two equal-sized groups of five individuals each, where the candidates in the first group are assigned scores $S_1 \sim \mathcal{U}(0,1)$, and in the second group - $S_2 \sim \mathcal{U}(0 + \delta, 1 + \delta)$, where $\delta = \{0.0, 0.1,\dots,0.9,1.0\}$. 
We sort the rankings according to the descending candidate scores and sample the Mallows distribution centered on the sorted rankings with difference values of $\theta$. We evaluate the Infeasible Index and NDCG of the samples.

\begin{figure}[ht!]
  \centering
  \includegraphics[width=0.4\textwidth]{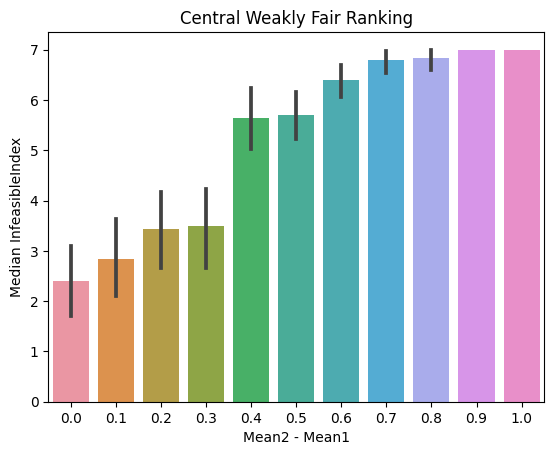}
  \caption{The Infeasible Index of the Central Ranking, as constructed by sampling from score distributions for each of the two groups (\Cref{sec:uniform-exp}). Specifically, the x-axis depicts the difference in means between the score distributions of the two groups. Confidence intervals were obtained via bootstrapping ($n=1000$).}
  \label{plot:experiment2initialfairness}
\end{figure}


\subsubsection{Experimental Results} 
\begin{figure}[tb!]
  \centering
  \includegraphics[width=0.45\textwidth]{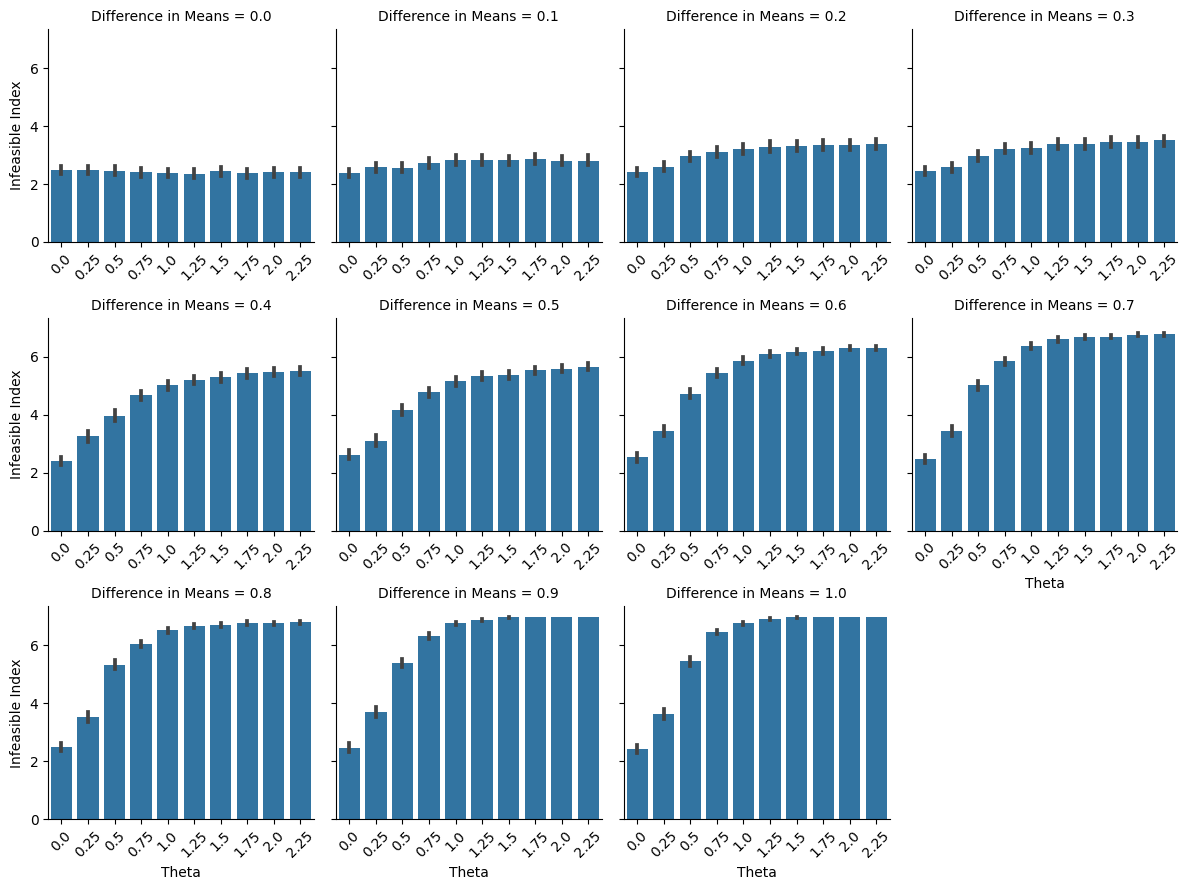}
  \caption{Mallow's distribution and Infeasible Index. Each subplot corresponds to a difference in means between the score distributions of the two groups.  We sample five individuals for each group, where the candidates in the first group are assigned scores $S_1 \sim \mathcal{U}(0,1)$, and in the second group - $S_2 \sim \mathcal{U}(0 + \delta, 1 + \delta)$, where $\delta$ is the difference in means. The subplots depict the mean value of the Infeasible Index of the samples from the Mallows distribution centered on the initial ranking. Confidence intervals were obtained via bootstrapping ($n=1000$).}
  \label{plot:experiment2infind}
\end{figure}

\begin{figure}[tb!]
  \centering
  \includegraphics[width=0.45\textwidth]{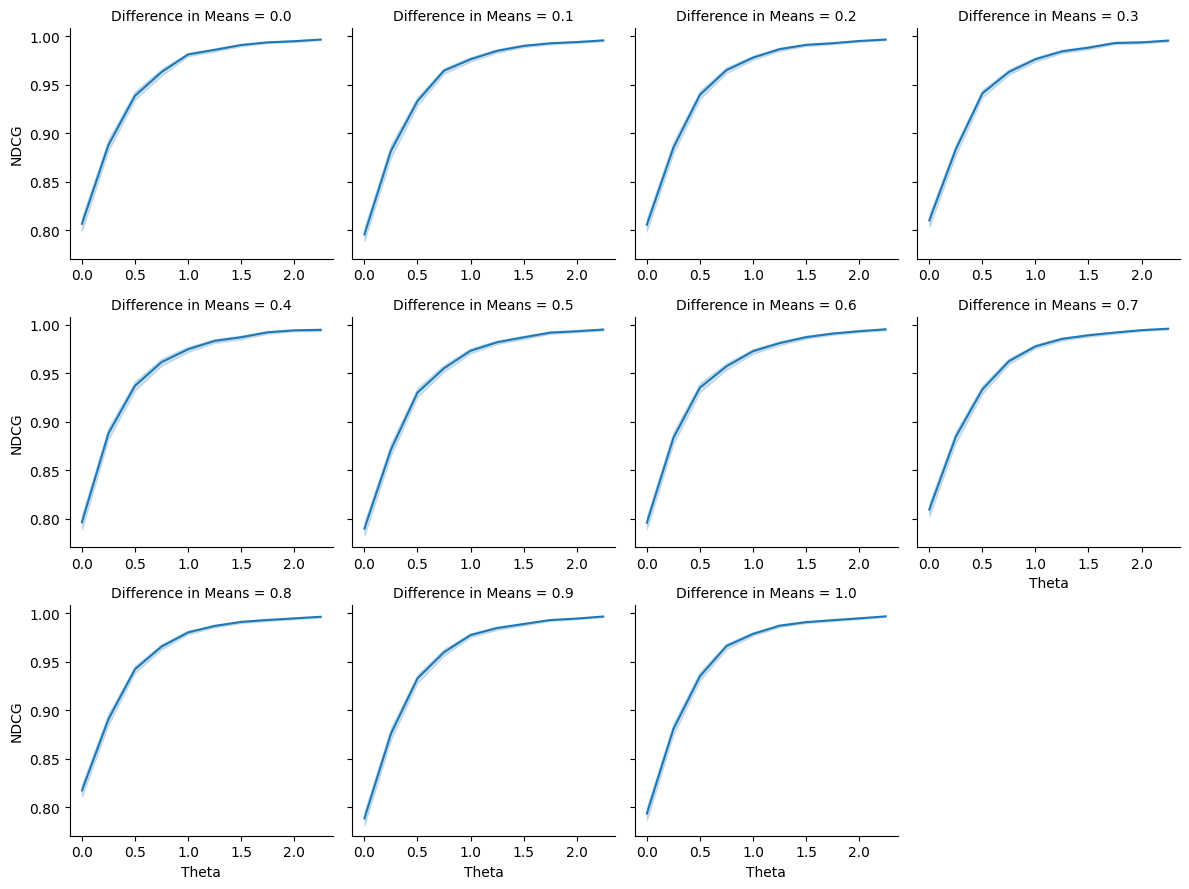}
  \caption{Mallow's distribution and NDCG. Each subplot corresponds to a difference in means between the score distributions of the two groups. We sample five individuals for each group, where the candidates in the first group are assigned scores $S_1 \sim \mathcal{U}(0,1)$, and in the second group - $S_2 \sim \mathcal{U}(0 + \delta, 1 + \delta)$, where $\delta$ is the difference in means. The subplots depict the mean value of NDCG of the samples from the Mallows distribution centered on the initial ranking. Confidence intervals were obtained via bootstrapping ($n=1000$).}
  \label{plot:experiment2ndcg}
\end{figure}

Figs. \ref{plot:experiment2infind} and \ref{plot:experiment2ndcg} depict the experimental results in terms of evaluating how the Infeasible Index and the NDCG change as we start from the corresponding Infeasible Indexes as depicted in a rankings  Fig. \ref{plot:experiment2initialfairness}. As for the Infeasible Index, we notice similar behavior as in the \Cref{sec:mallows-ii-exp}. We can see that as the dispersion parameter increases the NDCG converges to 1. We note that the NDCG of the central ranking is 1. Therefore we can conclude that there is a trade-off between the NDCG and the Infeasible Index while we sample from Mallow's distribution with different dispersion parameters. 

Figs. \ref{plot:experiment2infind} and \ref{plot:experiment2ndcg} depict the experimental results in terms of evaluating how the Infeasible Index and the NDCG change as we start from the corresponding Infeasible Indexes of the central rankings as depicted in Fig. \ref{plot:experiment2initialfairness}. As for the Infeasible index, we notice similar behavior as in \Cref{sec:mallows-ii-exp}. We can see that as the dispersion parameter increases, the Infeasible Index converges to the Infeasible Index of the Central Ranking and the NDCG converges to 1, which is the NDCG of the Central Ranking. Therefore, we can conclude that there is a trade-off between the NDCG and the Infeasible Index. 

\subsection{Experiments with the German Credit Dataset}
\label{sec:german-credit-exp}
In the third experiment, we utilized the German Credit dataset \cite{german_credit_statlog} to evaluate how well the Mallows randomization method performs in a practical scenario, where we have partial knowledge regarding some of the protected attributes and aim to evaluate the result in terms of an unknown protected attribute. We conduct a thorough experimental analysis with several state-of-the-art postprocessing algorithms that are designed to ensure fairness in rankings regarding specific attributes. To emulate real-world conditions, we introduce noise into their fairness constraints to simulate imperfect knowledge about group membership.

\subsubsection{Dataset Description}
We utilize the German Credit dataset \cite{german_credit_statlog}, following the work of \cite{yang2016measuring,dong2022rankaggregation}. For the ranking of the candidates, we use the Credit Amount attribute. 
We aggregate the binary attributes $Sex$ and $Age$ into the $Sex-Age$  protected attribute with four values and consider the information of this attribute as known, with little or no noise. We evaluate the fairness of the algorithms in terms of a third attribute, named $Housing$, with three values. We regarded the $Housing$ attribute as unknown; therefore, it could not be used as information for any algorithm.  The distribution of the groups defined by these attributes is shown in \Cref{tab:german_groups}.

     
   

\begin{table}[tb!]
  \centering
   \caption{Distribution of groups defined by Age, Sex, and Housing in the German Credit dataset.}
    \label{tab:german_groups}
\begin{tabular}{ccccc}
\hline
\multicolumn{1}{c}{\multirow{2}{*}{Age-Sex}} & \multicolumn{3}{c}{Housing}                                                     & \multicolumn{1}{l}{\multirow{2}{*}{Total}} \\ \cline{2-4}
\multicolumn{1}{c}{}                         & \multicolumn{1}{l}{free} & \multicolumn{1}{l}{own} & \multicolumn{1}{l}{rent} & \multicolumn{1}{l}{}                       \\ \hline
$<35$ - female                                 & 2                         & 131                      & 80                        & 213                                         \\
$<35$ - male                                   & 23                        & 261                      & 51                        & 335                                         \\
$\geq 35$ - female                             & 17                        & 65                       & 15                        & 97                                          \\
$\geq 35$ - male                               & 66                        & 256                      & 33                        & 355                                         \\ \hline
Total                                          & 108                       & 713                      & 179                       & 1000 \\ \hline                                       
\end{tabular}
\end{table}

\subsubsection{Experimental Setup}
We executed the state-of-the-art ApproxMultiValuedIPF \cite{dong2022rankaggregation}, DetConstSort \cite{Geyik_2019} as well as the ILP algorithm described in subsection \ref{sec:ilp}, using as input a weakly-p-fair ranking with respect to the combined $Sex-Age$ protected attribute. The algorithms were run in their vanilla version and with some noisy representation constraints on the combined $Sex-Age$ protected attribute. 
Specifically, we introduced noise into the calculations of the constraints by each of the aforementioned algorithms in the following ways:

\begin{itemize}
    \item ApproxMultiValuedIPF: we added an independent sample from $N(0,\sigma)$ to each of the weights at the weight calculation step ( Algorithm 2, line 2 of \cite{dong2022rankaggregation})
    \item DetConstSort: we added an independent sample from $N(0,\sigma)$ to each of tempMinCounts ( Algorithm 3, line 7 of \cite{Geyik_2019})
    \item ILP: given $X_{ij}, Y_{ij} \sim |N(0,\sigma)|$, we modified the calculation of constraints for each group $G_p$ such that:
    \begin{align*}
    & \lfloor \beta_p \ell \rfloor - X_{ij} \leq \sum_{i \in G_p} \sum_{j = 1}^\ell x_{ij}  \leq \lceil \alpha_p \ell \rceil + Y_{ij} & \forall \ell \in [k]
    \end{align*}
    This was done to lessen the probability of making the problem infeasible, while still retaining noise.
\end{itemize}

We repeated this step 15 times to reliably measure the effect of the noise.

We also run the Mallows algorithm on the same weakly-p-fair ranking, using the weakly-p-fair ranking as a central ranking and dispersion parameters of 0.5 and 1, taking 1 or the best of 15 samples.

We evaluate the fairness of the output rankings using the Infeasible Index with respect to the $Housing$ attribute and the utility of the output rankings using NDCG.
The experiment is repeated for rankings of size {10, 20, 30, 40, 50, 60, 70, 80, 90, 100}.

\begin{figure}[tb!]
  \centering
  \subfigure[$\theta = 0.5$, No noise]{\includegraphics[scale=0.23]{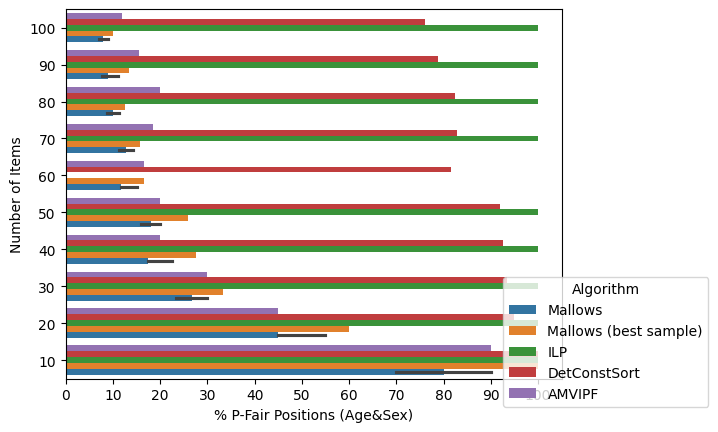}}\quad
  \subfigure[$\theta = 1$, No noise]{\includegraphics[scale=0.23]{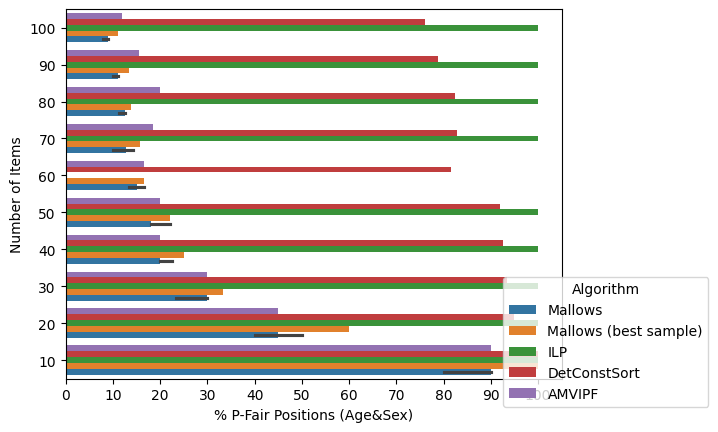}}\quad
  \subfigure[$\theta = 0.5, \sigma = 1$]{\includegraphics[scale=0.23]{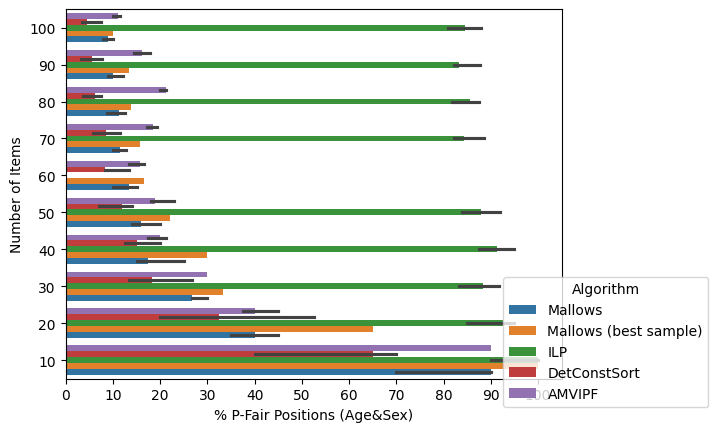}}\quad
  \subfigure[$\theta = 1, \sigma = 1$]{\includegraphics[scale=0.23]{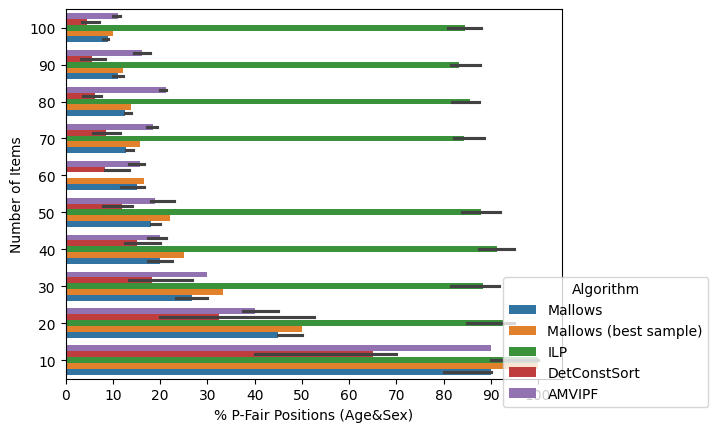}}\quad
  \caption{
  Rankings constructed with noisy representation constraints on the combined $Age-Sex$ protected attribute from an initial weakly-p-fair ranking with respect to the combined $Age-Sex$ protected attribute. The plots show the median percentage of positions satisfying P-fairness w.r.t. the $Age-Sex$ protected attribute. Confidence intervals were obtained via bootstrapping ($n=1000$). In Subfigure (a) the $\theta$ parameter of the Mallows distribution is set to $0.5$, and no noise is added to the constraints. In Subfigure (b) $\theta=1$ and no noise is added to the constraints. In Subfigure (c) $\theta=0.5$ and Gaussian noise $\xi\sim \mathcal{N}(0,1)$ is added to the constraints. In Subfigure  (d) $\theta=1$ and Gaussian noise $\xi\sim \mathcal{N}(0,1)$ is added to the constraints.  
  }
  \label{plot:pfair}
\end{figure}

\begin{figure}[tb!]
  \centering
  \subfigure[$\theta = 0.5$, No noise]{\includegraphics[scale=0.23]{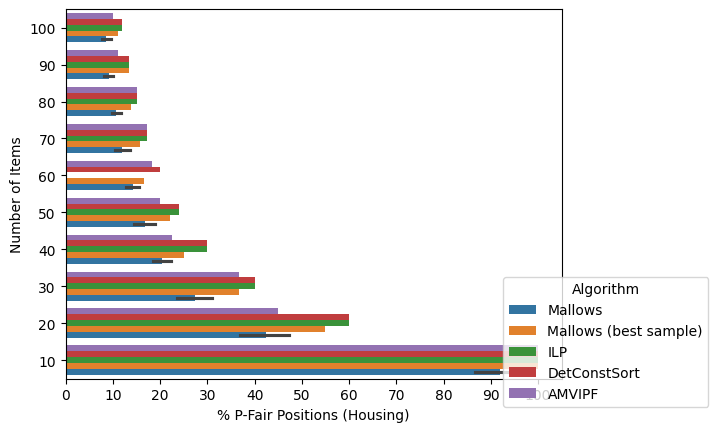}}\quad
  \subfigure[$\theta = 1$, No noise]{\includegraphics[scale=0.23]{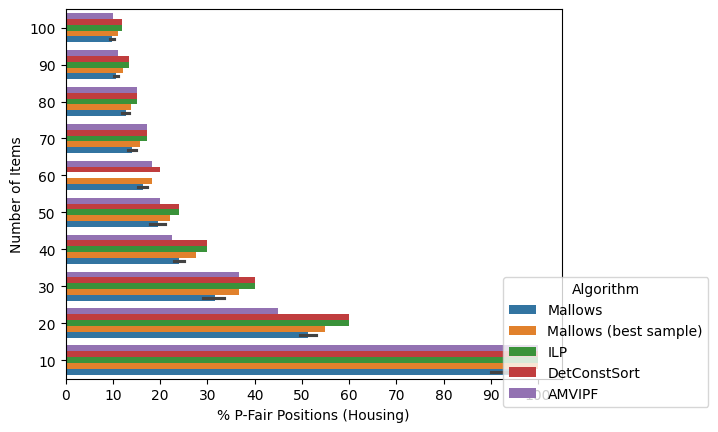}}\quad
  \subfigure[$\theta = 0.5, \sigma = 1$]{\includegraphics[scale=0.23]{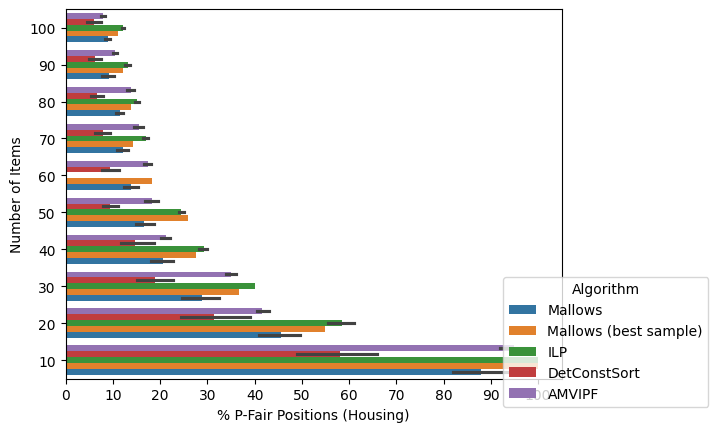}}\quad
  \subfigure[$\theta = 1, \sigma = 1$]{\includegraphics[scale=0.23]{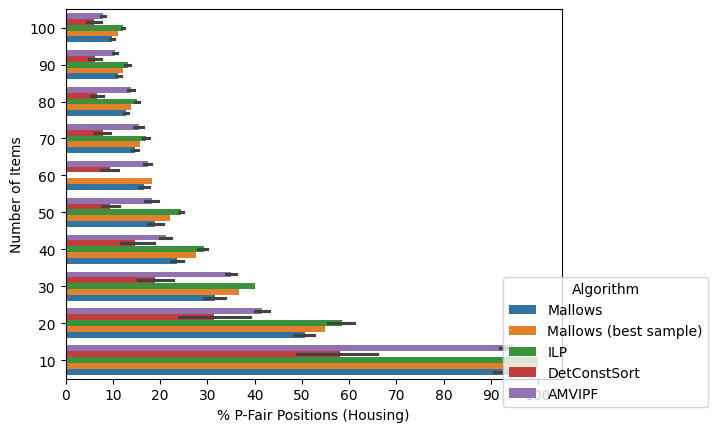}}\quad
  \caption{
  Rankings constructed with noisy representation constraints on the combined $Age-Sex$ protected attribute from an initial weakly-p-fair ranking with respect to the combined $Age-Sex$ protected attribute. The plots show the median percentage of positions satisfying P-fairness w.r.t. the $Housing$ protected attribute. Confidence intervals were obtained via bootstrapping ($n=1000$). In Subfigure (a) the $\theta$ parameter of the Mallows distribution is set to $0.5$, and no noise is added to the constraints. In Subfigure (b) $\theta=1$ and no noise is added to the constraints. In Subfigure (c) $\theta=0.5$ and Gaussian noise $\xi\sim \mathcal{N}(0,1)$ is added to the constraints. In Subfigure  (d) $\theta=1$ and Gaussian noise $\xi\sim \mathcal{N}(0,1)$ is added to the constraints.  
  }
  \label{plot:pfair_hous}
\end{figure}

\begin{figure}[tb!]
  \centering
  \subfigure[$\theta = 0.5$, No noise]{\includegraphics[scale=0.2]{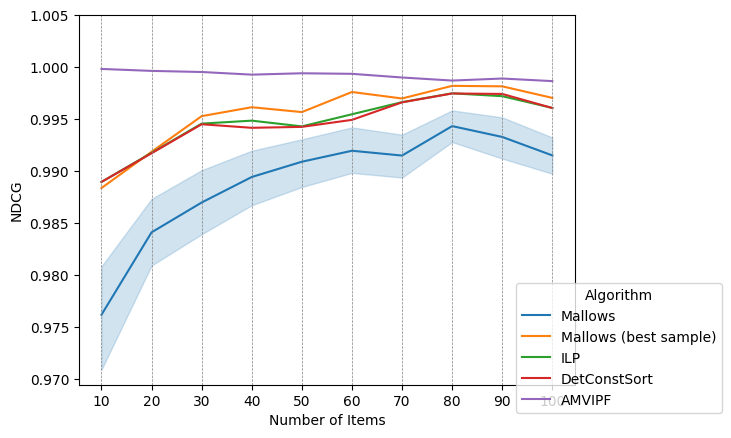}}\quad
  \subfigure[$\theta = 1$, No noise]{\includegraphics[scale=0.2]{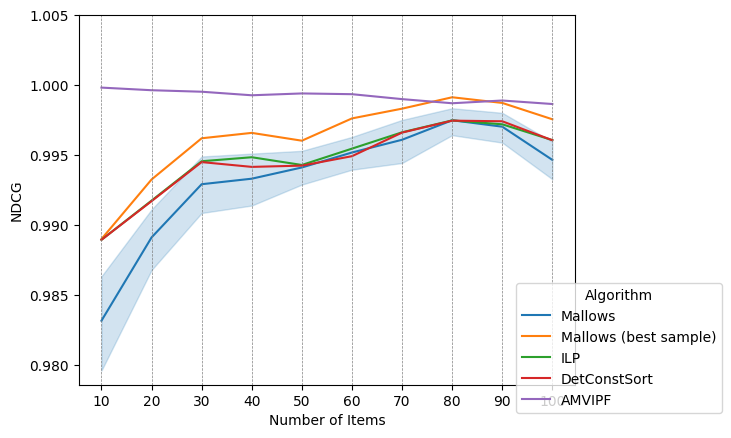}}\quad
  \subfigure[$\theta = 0.5, \sigma = 1$]{\includegraphics[scale=0.2]{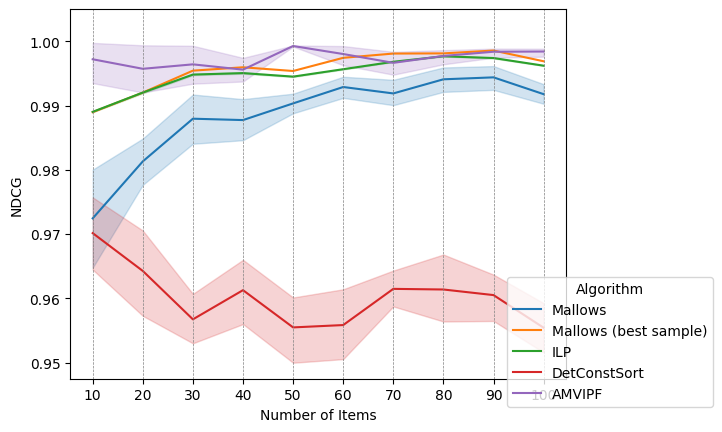}}\quad
  \subfigure[$\theta = 1, \sigma = 1$]{\includegraphics[scale=0.2]{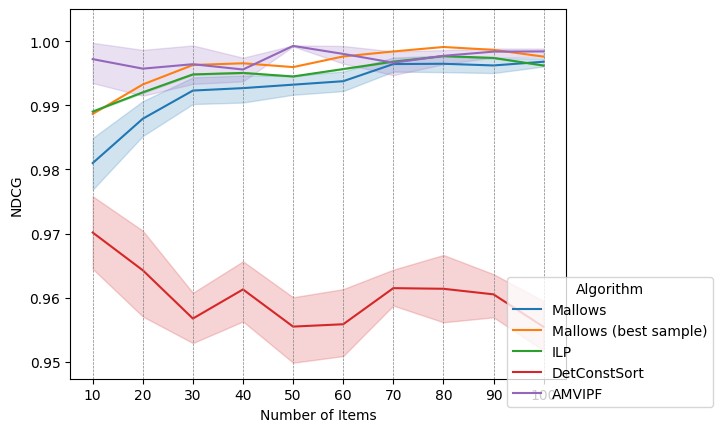}}\quad
  \caption{
  Mean NDCG (solid line) and $\pm1$ standard deviations away from the mean (shaded region) of the output rankings. Confidence intervals were obtained via bootstrapping ($n=1000$). In Subfigure (a) the $\theta$ parameter of the Mallows distribution is set to $0.5$, and no noise is added to the constraints. In Subfigure (b) $\theta=1$ and no noise is added to the constraints. In Subfigure (c) $\theta=0.5$ and Gaussian noise $\xi\sim \mathcal{N}(0,1)$ is added to the constraints. In Subfigure  (d) $\theta=1$ and Gaussian noise $\xi\sim \mathcal{N}(0,1)$ is added to the constraints.  }
  \label{plot:ndcg}
\end{figure}

\subsubsection{Experimental Results}

The experimental results are presented in Figs.~\ref{plot:pfair_hous} and \ref{plot:ndcg}. 
Fig.~\ref{plot:pfair_hous} shows the median percentage of positions satisfying P-fairness with respect to the $Housing$ protected attribute. Since DetConstSort,  ApproxMultiValuedIPF and the ILP use the $Age-Sex$ protected attribute to create the fair ranking, the resulting ranking fairness has to do with another attribute's distribution, therefore we cannot have any guarantees.  For comparison only, we include Fig. \ref{plot:pfair}, so that it is clear how the same ranking can have different fainress scores according to different attribute. We argue that the addition of noise can improve the results regarding different protected attributes, leading to a more balanced output, regardless of the target protected attribute. This approach seems like a compromise among all the different protected attributes that may be present and about which we have no knowledge.


As shown in Fig.~\ref{plot:pfair}, under noisy conditions, the Mallows algorithm performs well compared to state-of-the-art algorithms (DetConstSort and ApproxMultiValuedIPF).
Fig.~\ref{plot:ndcg} shows the mean NDCG (as a solid line) and $\pm1$ standard deviations (shaded region). We can notice that as the number of items increase, the NDCG score increases for \Cref{algo1}, and the performance of the best sample from \Cref{algo1} approaches the NDCG curve for the ILP.

\section{Conclusions}

We have introduced a randomized algorithm for post-processing rankings, demonstrating its efficacy in balancing fairness across multiple attributes through the strategic use of noise.  Our findings highlight the potential of noise injection to address fairness concerns, particularly in scenarios where individual attribute information is limited or unavailable.

As future work, we propose developing a systematic methodology for incorporating noise into rankings.
This could involve exploring various ``noise distributions'' or tuning parameters within the noise distribution, such as dispersion in the case of Mallow's model. 
This approach aims to provide fairness guarantees by leveraging knowledge about the distribution of multiple attributes while respecting individual attribute privacy. Such advancements will contribute significantly to the field of fairness-aware ranking algorithms and their practical applications in real-world settings.



\FloatBarrier
\section*{Acknowledgment}

This research has been supported by European Union’s Horizon Europe research and innovation programme under grant agreement No. GA 101070568 
(Human-compatible AI with guarantees).

\bibliographystyle{unsrt}
\bibliography{refs}

\end{document}